\documentclass{ctexart}
\usepackage{amsmath}
\usepackage{booktabs} 
\usepackage{graphicx} 
\usepackage{caption}
\usepackage{longtable}
\usepackage{booktabs}
\usepackage{array}
\usepackage{amsmath,amssymb}
\usepackage{float}

\newcolumntype{L}[1]{>{\raggedright\arraybackslash}p{#1}}
\newcolumntype{R}[1]{>{\raggedleft\arraybackslash}p{#1}}

\usepackage{array}
\newcolumntype{L}[1]{>{\raggedright\arraybackslash}p{#1}}
\newcolumntype{R}[1]{>{\raggedleft\arraybackslash}p{#1}}

\usepackage{fancyhdr}

\fancypagestyle{plain}{
  \fancyhead{}
  
}
\fancypagestyle{firstpage}{
  \fancyhead{}                      

  \fancyfoot[L]{\small *Correspondence to: suiyicheng@nankai.edu.cn}
  \fancyfoot[R]{\small \thepage} 
}

\usepackage[
    backend=biber,
    style=numeric-comp,
    sorting=none,
    hyperref=true
]{biblatex}
\title{\Large \textbf{A Two-Stage GPU Kernel Tuner Combining Semantic Refactoring and Search-Based Optimization}}
\author{Qiuyi Qu\textsuperscript{1，2}, 
Yicheng Sui\textsuperscript{1*},
Yufei Sun\textsuperscript{1},
Rui Chen\textsuperscript{1},
Xiaofei Zhang\textsuperscript{2},\\
Yuzhi Zhang\textsuperscript{1},
Haofeng Wang\textsuperscript{1},
Ge Lan\textsuperscript{1},
Ning Zhang\textsuperscript{2},
Manfei Zhao\textsuperscript{2}\\
{\small $^{1}$NanKai University}     {\small $^{2}$Beijing Institute of Computer Technology and Applications}}

\date{}

\begin{document}
\maketitle

\thispagestyle{firstpage} 

\noindent\textbf{Abstract:}\hspace{0.5em}
GPU code optimization is a key bottleneck for high-performance computing (HPC) and for large-model training and inference. Expert-written kernels require semantics-level refactoring and fine-grained parameter tuning, which is costly, slow, and hard to scale. Although LLM-based agents have reduced the barrier to kernel generation and optimization, most existing approaches directly rewrite code, leaving critical performance choices implicit and lacking controllable parameterized interfaces and systematic search; consequently, speedups can be unstable and the optimization process is difficult to interpret and reproduce. We propose a two-level GPU code tuner that couples semantic refactoring with search-based optimization: within an agentic iterative feedback loop, candidate kernels are explicitly rewritten into parameterizable templates, and template parameters are tuned via guided search under hardware constraints using profiling signals. We evaluate on real-world kernels extracted from SGLang, where our method achieves over 3× speedup versus native SGLang implementations on a set of kernels. Compared with agent-only direct rewriting, templating plus search reduces iteration randomness, improves interpretability and reproducibility, and more systematically approaches high-performance configurations. The framework can be extended to other backends such as OpenCL and HIP to provide automated optimization for practical service workloads.

\section{Introduction}
As AI workloads rapidly evolve in scale and structure, GPU kernel performance has increasingly become a critical bottleneck for training and inference system efficiency. Although large language models have demonstrated strong capabilities in code generation and program rewriting, consistently producing GPU kernels that are compilable, correct, and close to expert-level performance remains challenging. The root cause lies in the highly combinatorial optimization space and the tight coupling between structural transformations and hardware resource constraints. Free-form source-level rewriting is therefore difficult to control and reproduce, and can easily introduce semantic deviations and performance variability.

To address this issue, we leverage the automation capability of multi-agent iterative optimization and develop a unified formulation that integrates semantics-preserving refactoring with resource-feasible search. Based on this formulation, we design a two-level GPU code tuner that combines semantic refactoring and parameter search. Specifically, guided by measurement feedback and hardware constraints, we first refactor candidate kernels into explicit parameterizable templates, so that optimization intent is mapped into a manageable and reproducible parameter space. We then perform search-based fine-grained tuning within the resource-feasible region, closing the loop through compilation/execution, correctness verification, and performance evaluation. Furthermore, we use measurement feedback to drive semantics-level structural rewriting, continuously raising the reachable performance upper bound and thereby stably approaching high-performance implementations while preserving correctness.

Our contributions are as follows:

(1) We build an automated optimization framework for GPU kernels that unifies template-based refactoring and search-based fine-grained tuning within a single pipeline. By combining hardware resource awareness with measurement feedback, the optimization process is transformed from uncontrolled free-form rewriting into a reproducible, constraint-aware closed loop of parameterized search and validation.

(2) We design a semantics–parameter hierarchical optimization mechanism. At the semantic level, semantics-preserving rewriting improves the structural performance upper bound; at the parameter level, resource-constraint derivation and search-based optimization drive the kernel toward near-optimal mappings, yielding stable performance gains under correctness gating.

(3) We implement an end-to-end closed-loop tuner with multi-agent collaboration, covering semantics-level refactoring, templating, feasible-region derivation, tuning search, correctness verification, and performance measurement. Measurement signals further guide subsequent semantic rewriting, enabling a traceable and reproducible iterative optimization process.

\section{Related Work}

\textbf{Multi-Agent Systems (MAS).} MAS consists of multiple interacting agents that collaborate to solve complex shared tasks beyond the capability of a single agent. This paradigm naturally aligns with programming-oriented workflows: complex software development processes can be decomposed into subtasks such as planning, implementation, testing, and profiling, which can then be assigned to different agents for coordinated and iterative progress. In recent years, a growing body of work has explored multi-agent frameworks (e.g., AutoGen, Trace, MetaGPT), demonstrating strong capability and empirical effectiveness on benchmark tasks including mathematical reasoning and code generation~\cite{wu2024autogen,cheng2024trace,hong2023metagpt,li2023camel,huang2023agentcoder,qian2024chatdev,wei2024agentinterfaces,yang2024sweagent}. However, systematic application of MAS to parallel code optimization remains relatively limited. A fundamental reason is that performance optimization for modern accelerators relies heavily on specialized knowledge of hardware microarchitecture, parallel execution models, and memory hierarchies, with complex optimization objectives and constraints. As a result, while multi-agent approaches are promising, they also introduce new challenges in modeling and coordination.

\textbf{Compiler- and learning-based methods for GPU kernel optimization.} For decades, parallel code optimization has been driven primarily by compiler frameworks and domain-specific languages (DSLs). Representative systems include Halide, TVM, MLIR, TensorFlow XLA, and NVIDIA CUTLASS~\cite{ragankelley2013halide,lattner2020mlir,girija2016tensorflow,thakkar2023cutlass,rotem2018glow,weirichard2017dlvm,jia2019beyonddatamodel,zheng2021neoflow,wei2025taskbasedamr}. These systems provide high-level abstractions for expressing tensor computations and leverage compiler optimization pipelines to perform scheduling transformations, operator fusion, and memory-hierarchy mapping, thereby reducing the cost of manual optimization.

To further approach expert-level performance, autotuning has become a key technique. Frameworks such as AutoTVM, Ansor, AMOS, KernelTuner, and OpenTuner explore large optimization spaces via search and machine learning to identify hardware-specialized configurations~\cite{chen2018tvm,zheng2020ansor,zheng2022amos,vanwerkhoven2019kerneltuner,ansel2014opentuner}. Building on this foundation, recent systems (e.g., Triton, Mirage, ThunderKittens) further expand the design space of the ``DSL + tuning'' paradigm. For instance, Triton introduces a tile-level intermediate representation coupled with autotuning, enabling generated kernels to reach performance close to hand-optimized implementations under certain conditions~\cite{tillet2019triton,wu2025mirage,spector2025thunderkittens}. Despite this progress, compiler/DSL-driven approaches still face two prominent limitations. First, without substantial tuning effort, they often struggle to consistently achieve expert-level performance, and their generalization across hardware platforms remains insufficient\cite{spector2025thunderkittens}. Second, these systems typically rely on relatively rigid compilation pipelines, which incur high development and maintenance costs and demand significant engineering investment. In other words, there remains a clear tension between \emph{high-performance attainability} and \emph{system scalability/portability}.

\textbf{LLM-driven high-performance code generation and optimization.} Early work on large language model (LLM)-based code generation primarily targeted general-purpose programming tasks (e.g., AlphaCode) and demonstrated promising potential\cite{li2022alphacode}. In recent years, the focus has increasingly shifted toward performance-oriented, domain-specific code generation, spanning vectorization\cite{taneja2025llmvectorizer,zheng2025vectrans}, assembly-level optimization\cite{wei2025assemblyrl}, parallel program generation for DSLs\cite{wei2025agentinterfaces,wei2024agentinterfaces,wei2025mapple}, and tensor program optimization\cite{zhai2024tensorlanguage}. Among these directions, automatic generation of high-performance GPU kernels has emerged as a particularly active line of research\cite{ouyang2025kernelbench,li2025tritonbench}. Unlike general-purpose code generation, performance optimization offers verifiable reward signals (e.g., latency, throughput, and bandwidth utilization). Consequently, an iterative optimization paradigm of ``generate--evaluate--improve'' has become increasingly prevalent: the model first produces candidate kernels, which are then refined through a feedback loop involving compilation checks, correctness verification, runtime profiling, or self-reflection\cite{brauckmann2025tensorize,wei2024agentinterfaces,chen2025deepseekr1,wang2025geak,wei2025astra}. 
 However, existing LLM- or multi-agent-based tuning processes largely rely on LLM-driven feedback loops and lack an interpretable and controllable mechanism for fine-grained parameter tuning. This often causes key parameter boundaries and interaction effects under hardware resource constraints to be missed, leading to incomplete exploration and high variance in performance.

Overall, prior work on parallel code generation and optimization exhibits fragmented capabilities and insufficient closed-loop integration, making it difficult to simultaneously achieve controllability, reproducibility, and expert-level performance. Targeting the common bottleneck that high-performance parallel implementations for intelligent computing systems are difficult to obtain automatically and reliably, this study treats resource-aware autotuning as a key lever for approaching expert implementations. We aim to build an end-to-end tuning closed loop with multi-agent collaboration, modularly chaining hardware awareness, parameter space derivation, compilation and correctness verification, performance evaluation, and optimization updates into a sustainable iterative optimization pipeline. The resulting framework is intended to uniformly support both CUDA and OpenCL backends, enable cross-platform performance transfer, and stably produce near-expert high-performance parallel implementations across diverse workloads.

\section{Method}
\subsection{Task Definition}
The goal of GPU code tuning is to generate an optimized parameterizable templated kernel and to search for the best general configuration and specialized configuration, so that it runs faster than the original GPU kernel while preserving functional correctness. We formalize the task as follows.

\paragraph{Correctness}
Let the input domain be $X$ and the output space be $Y$. The original kernel is
\[
S: X \rightarrow Y.
\]
Ideally, the kernel $\tilde{S}$ produced by the two-level optimization should be equivalent to the original kernel over the entire input domain:
\[
\forall x \in X:\ \tilde{S}(x) = S(x).
\]
Considering floating-point errors, we allow approximate equivalence under a metric $d$ and a tolerance $\varepsilon \ge 0$:
\[
\forall x \in X:\ d(\tilde{S}(x), S(x)) \le \varepsilon.
\]
In practice, correctness is verified on a finite test set $T = \{x_i\}_{i=1}^{m}$:
\[
\max_{1\le i\le m}\ d(\tilde{S}(x_i), S(x_i)) \le \varepsilon.
\]

\paragraph{Performance Measurement and Evaluation}
Let $T_h(S,x)$ denote the runtime of kernel $S$ on hardware $h$ for processing input $x$.
For each input $x\in T$, we focus on the best achievable performance over the feasible parameter set:
\[
T_h^*(x)=\min_{\theta \in \Theta_{\mathrm{feasible}}}\ T_h\!\left(\mathrm{Instantiate}(\tilde{S},\theta),\,x\right).
\]
Here, $\Theta_{\mathrm{feasible}}$ denotes the set of parameter assignments that satisfy both correctness and hardware resource constraints.
We define the speedup over the original kernel $S$ as
\[
\sigma^*(x)=\frac{T_h(S,x)}{T_h^*(x)}
=\max_{\theta \in \Theta_{\mathrm{feasible}}}\ \frac{T_h(S,x)}{T_h\!\left(\mathrm{Instantiate}(\tilde{S},\theta),\,x\right)}.
\]
To aggregate results over the test set $T$, we use the geometric mean speedup:
\[
\sigma_T^*=\left(\prod_{i=1}^{m}\sigma^*(x_i)\right)^{1/m}.
\]
Here, $m$ denotes the number of samples in the test set $T$.
The optimization objective is to maximize this speedup while preserving correctness.

\subsection{Multi-agent System}

\begin{figure}[H]
    \centering
    \includegraphics[width=\textwidth]{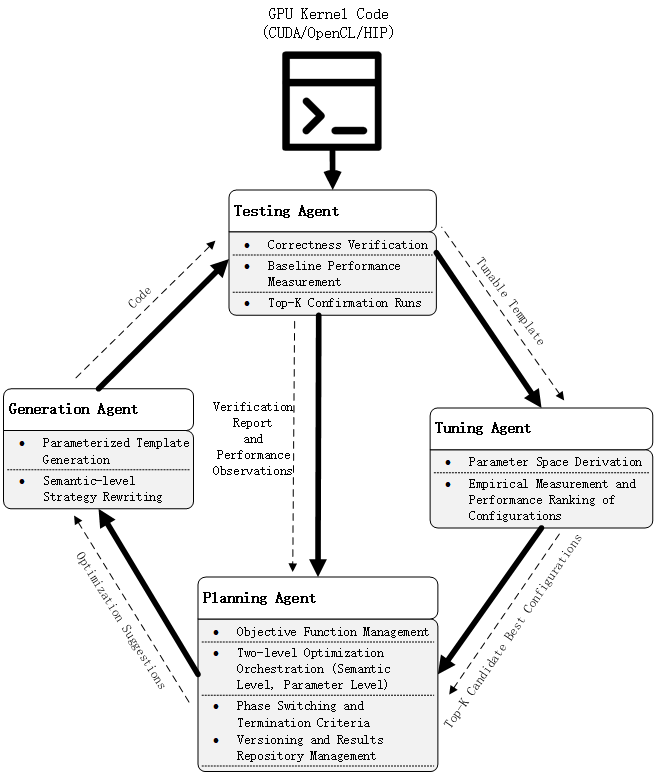}
    \caption{A two-level GPU code tuning agentic pipeline combining semantic refactoring and search-based optimization.}
    \label{fig:gpu_optimize-en}
\end{figure}

We build an iterative closed-loop optimization workflow around four agents---a generation agent, a tuning agent, a testing agent, and a planning agent. Figure~\ref{fig:gpu_optimize-en} illustrates the four-agent collaborative framework for two-level optimization of GPU kernels. The framework takes existing CUDA/OpenCL/HIP kernels as input and realizes a layered closed loop that combines semantic-level strategy rewriting with parameter-level search-based tuning, thereby enforcing correctness constraints, enabling performance evaluation, and driving iterative convergence of candidate implementations. In the overall workflow, the planning agent orchestrates global objectives and iteration scheduling; the generation agent constructs tunable kernel families at both the semantic and template levels; the tuning agent performs empirical configuration ranking through on-device measurements within the constraint-feasible region; and the testing agent provides unified correctness verification and baseline comparison, feeding verification reports and performance observations back to drive the next optimization round.

\textbf{Planning agent.} The planning agent orchestrates and governs the global optimization process. It centrally manages the objective function and budget constraints, coordinates switching and termination criteria between semantic-level and parameter-level optimization, and maintains a versioned repository of artifacts and results to ensure that the iterative process is traceable and reproducible.

\textbf{Generation agent.} The generation agent is responsible for the semantic-level and template-based stages of the two-level optimization. At the semantic level, it performs strategy-family rewriting to form structured candidate directions. At the template level, it refactors the kernel into a parameterized implementation family and explicitly exposes tunable options, providing interpretable parameter interfaces and constraint rationales for subsequent search.

\textbf{Tuning agent.} The tuning agent targets parameter-level optimization. Under hardware resource and problem-size constraints, it derives and prunes the search space and performs search-based autotuning. It compiles and runs candidate configurations and measures their true performance, producing empirical results and a performance ranking, and submits the Top-$K$ candidates for subsequent re-testing and convergence decisions.

\textbf{Testing agent.} The testing agent provides a unified verification and evaluation benchmark. It is responsible for correctness verification of candidate kernels, baseline performance measurement, and confirmation re-tests of Top-$K$ candidates. Its verification reports and performance observations serve as reliable feedback for planning and tuning, enabling search-strategy updates, triggering semantic-level rewriting, or determining the final best implementation.

Algorithm~1 presents a two-level GPU tuning pipeline driven by four collaborative agents. The procedure first constructs a finite test set $T$ from the baseline kernel $S_0$ and fixes $S_{\mathrm{ref}}=S_0$ as the semantic reference. Over $R$ iterations, the planning agent generates semantic-level suggestions $g$ based on measurement feedback, and the generation agent performs semantics-preserving rewriting and templates the kernel into $\tilde{S}_r$. The testing agent applies correctness gating under tolerance $\epsilon$ to filter out candidates that violate semantic constraints. For templates that pass the gate, the tuning agent derives a feasible parameter space $\theta_r^{\mathrm{feasible}}$ under the hardware resource limits $r(h)$, searches configurations to minimize runtime, and produces a deployable general configuration $\theta_r^{\mathrm{gen}}$ and a specialized configuration table $D_r^{\mathrm{spec}}$. The system logs results from all rounds and returns the globally optimal $(\tilde{S}^*, \theta^{\mathrm{gen}*}, D^{\mathrm{spec}*})$, enabling traceable and reproducible optimization.\\
{\renewcommand{\tablename}{Algorithm}%
\renewcommand{\arraystretch}{1.05}%
\begin{longtable}{@{}R{0.06\linewidth}L{0.94\linewidth}@{}}
\specialrule{1.2pt}{0pt}{0pt} 
\caption{A Two-Level GPU Code Tuner: Semantic Refactoring and Search-Based Optimization}
\label{alg:two_level_gpu_tuner}\\
\specialrule{0.6pt}{0pt}{0pt} 

\multicolumn{2}{@{}l@{}}{\textbf{Input:}}\\[-2pt]
 & Baseline GPU kernel $S_0:\mathcal{X}\rightarrow\mathcal{Y}$\\
 & Target hardware $h$ and resource upper-bound vector $r(h)$\\
 & Distance metric $d(\cdot,\cdot)$ and tolerance $\epsilon\ge 0$\\
 & Searchable parameter vector $\theta\in\Theta$ (discrete candidate space)\\
 & Number of semantic rounds $R$\\[2pt]

\multicolumn{2}{@{}l@{}}{\textbf{Define:}}\\[-2pt]
 & \texttt{TestingAgent}\;\(\triangleright\)\;Generate or run tests; validate correctness; measure runtime; collect performance signals\\
 & \texttt{GenerationAgent}\;\(\triangleright\)\;Apply semantics-preserving rewriting; produce a parameterizable template\\
 & \texttt{TuningAgent}\;\(\triangleright\)\;Derive/prune the feasible parameter space; search/rank configurations; output deployable configurations\\
 & \texttt{PlanningAgent}\;\(\triangleright\)\;Propose semantic-level suggestions from measurement signals; schedule rounds/budgets; select across rounds\\
 & $\mathit{Log}$\;\(\triangleright\)\;List of $(\textit{round},\,\textit{code},\,\textit{pass},\,\textit{performance},\,\textit{signals})$ for all iterations\\[2pt]

\multicolumn{2}{@{}l@{}}{\textbf{Output:}}\\[-2pt]
 & Best template $\tilde{S}^*$, best general configuration $\theta^{\mathrm{gen}*}$, best specialized table $D^{\mathrm{spec}*}$, and $\mathit{Log}$\\[4pt]

\multicolumn{2}{@{}l@{}}{\textbf{Procedure:}}\\[-2pt]
1:  & $T \leftarrow \texttt{TestingAgent}.\mathrm{GenerateTests}(S_0)$\;\(\triangleright\)\;Initialization: construct a finite test set\\
2:  & $S_{\mathrm{ref}} \leftarrow S_0$\;\(\triangleright\)\;Fix the semantic reference\\
3:  & $S_{\mathrm{prev}} \leftarrow S_0$;\ $m_{\mathrm{prev}} \leftarrow \emptyset$\;\(\triangleright\)\;Initialize previous code and signals\\
4:  & $\mathit{Log} \leftarrow [\,]$\;\(\triangleright\)\;Initialize the log\\
5:  & $(\tilde{S}^*,\theta^{\mathrm{gen}*},D^{\mathrm{spec}*}) \leftarrow \mathrm{None}$;\ $score^* \leftarrow +\infty$\;\(\triangleright\)\;Initialize the incumbent best\\

6:  & \textbf{for} $r \leftarrow 0$ \textbf{to} $R-1$ \textbf{do}\;\(\triangleright\)\;Iterative optimization starts\\
7:  & \quad \textit{Level 1: semantic refactoring}\\
8:  & \quad \textbf{if} $r>0$ \textbf{then}\\
9:  & \quad\quad $g \leftarrow \texttt{PlanningAgent}.\mathrm{Suggest}(m_{\mathrm{prev}})$\;\(\triangleright\)\;Propose semantic-level suggestions\\
10: & \quad\quad $S_r \leftarrow \texttt{GenerationAgent}.\mathrm{Rewrite}(S_{\mathrm{prev}},g)$\;\(\triangleright\)\;Semantics-preserving rewriting\\
11: & \quad \textbf{else}\;\ $S_r \leftarrow S_{\mathrm{prev}}$\\
12: & \quad \textbf{end if}\\

13: & \quad $pass_r \leftarrow \texttt{TestingAgent}.\mathrm{Validate}(S_r,S_{\mathrm{ref}},T,d,\epsilon)$\;\(\triangleright\)\;Correctness gating\\
14: & \quad \textbf{if} $pass_r=\mathrm{False}$ \textbf{then}\;\ $S_r \leftarrow S_{\mathrm{prev}}$\;\(\triangleright\)\;Rollback\\
15: & \quad \textbf{end if}\\[2pt]

16: & \quad \textit{Level 2: parameter search within the resource-feasible region}\\
17: & \quad $\tilde{S}_r \leftarrow \texttt{GenerationAgent}.\mathrm{Templatize}(S_r,\theta)$\;\(\triangleright\)\;Produce a parameterizable template\\
18: & \quad $\Theta_r^{\mathrm{feasible}} \leftarrow \texttt{TuningAgent}.\mathrm{DeriveFeasibleSpace}(\tilde{S}_r,h,r(h))$\;\(\triangleright\)\;Respect resource limits\\

19: & \quad \textbf{for each} $x\in T$ \textbf{do}\\
20: & \quad\quad $\theta_r^*(x) \leftarrow \texttt{TuningAgent}.\mathrm{Search}(\tilde{S}_r,x,\Theta_r^{\mathrm{feasible}})$\;\(\triangleright\)\;Minimize runtime under constraints\\
21: & \quad \textbf{end for}\\

22: & \quad $\theta_r^{\mathrm{gen}} \leftarrow \texttt{TuningAgent}.\mathrm{SelectGeneral}(\{\theta_r^*(x)\}_{x\in T},\tilde{S}_r,T)$\\
23: & \quad $D_r^{\mathrm{spec}} \leftarrow \texttt{TuningAgent}.\mathrm{BuildSpecialized}(\{\theta_r^*(x)\}_{x\in T},T)$\\

24: & \quad $perf_r \leftarrow \texttt{TestingAgent}.\mathrm{EvaluatePerformance}(\tilde{S}_r,\theta_r^{\mathrm{gen}},D_r^{\mathrm{spec}},S_{\mathrm{ref}},T,h)$\\
25: & \quad $m_r \leftarrow \texttt{TestingAgent}.\mathrm{CollectSignals}(\tilde{S}_r,\theta_r^{\mathrm{gen}},D_r^{\mathrm{spec}},T,h)$\\

26: & \quad $\mathrm{Append}(\mathit{Log},(r,S_r,\tilde{S}_r,\theta_r^{\mathrm{gen}},D_r^{\mathrm{spec}},pass_r,perf_r,m_r))$\\
27: & \quad $score_r \leftarrow \texttt{PlanningAgent}.\mathrm{Score}(perf_r)$\;\(\triangleright\)\;Negative geometric-mean speedup\\
28: & \quad \textbf{if} $score_r<score^*$ \textbf{then}\\
29: & \quad\quad $(\tilde{S}^*,\theta^{\mathrm{gen}*},D^{\mathrm{spec}*}) \leftarrow (\tilde{S}_r,\theta_r^{\mathrm{gen}},D_r^{\mathrm{spec}})$;\ $score^* \leftarrow score_r$\\
30: & \quad \textbf{end if}\\

31: & \quad $S_{\mathrm{prev}} \leftarrow S_r$;\ $m_{\mathrm{prev}} \leftarrow m_r$\;\(\triangleright\)\;Update state\\
32: & \textbf{end for}\\
33: & \textbf{return} $(\tilde{S}^*,\theta^{\mathrm{gen}*},D^{\mathrm{spec}*},\mathit{Log})$\\

\end{longtable}}
\addtocounter{table}{-1}

\subsection{Tuning parameter selection and search space derivation}
Parameterized templates require the generation agent to infer the tunable parameterization and construct template kernels according to kernel structure under semantics-preserving constraints. Since kernels differ in computation patterns, memory-access organization, and synchronization/reduction structures, the candidate parameters and tuning dimensions also vary accordingly. To balance search cost against the attainable performance ceiling, we select only a small set of key dimensions that offer the most broadly applicable benefits. The generation agent parameterizes execution strategies only, including thread-level parallel granularity, loop scheduling and unrolling, vectorization and alignment strategies, and the organization of shared memory and registers. It then derives parameter ranges and discrete candidate sets based on the target hardware resource limits, pruning combinations that would lead to compilation failures, invalid launches, or obvious inefficiency. In this way, the framework achieves stable and interpretable performance gains with a controlled tuning budget.

\section{Experimental Setup}

\textbf{Evaluation Metrics.}
Our evaluation considers both correctness and performance.
For correctness, we construct a test suite $\mathcal{T}$ covering different data types and input shapes.
For each test case, we compare the output of the final kernel produced by our method against the original SGLang implementation element-wise under identical inputs, and deem it correct if the mismatch is within a given tolerance.
For performance, we measure the runtime of the native SGLang , the baseline method Astra~\cite{wei2025astra}, and ours, and use the speedup over SGLang as the primary metric.

\textbf{Kernels.}
We select three representative CUDA kernels from the LLM serving framework SGLang~\cite{wei2025astra} as evaluation targets, corresponding to \texttt{silu\_and\_mul}, \texttt{fused\_add\_rmsnorm}, and \texttt{merge\_attn\_states}, which cover representative workload characteristics across different inference stages.
Experiments are conducted on multiple data types and input shapes:
\begin{enumerate}
  \item[(1)] Kernel1:
  We test 5 shapes $(n,h,d)$ under \texttt{fp16}/\texttt{bf16}:
  $(512,32,256)$, $(512,40,128)$, $(768,32,256)$, $(768,40,128)$, $(512,64,128)$.

  \item[(2)] Kernel2:
  We test 8 shapes $(m,n)$ under \texttt{fp32}:
  $(128,4096)$, $(256,4096)$, $(1024,4096)$, $(2048,8192)$,
  $(128,11008)$, $(256,13824)$, $(512,14336)$, $(1024,8192)$.

  \item[(3)] Kernel3:
  We test 5 shapes $(m,n)$ under \texttt{fp16}/\texttt{bf16}:
  $(16,4096)$, $(32,5120)$, $(32,8192)$, $(16,12288)$, $(64,8192)$.
\end{enumerate}

\textbf{Performance Measurement.}
We report two types of results on representative input shapes: the best specialized configuration per shape, and the best general configuration that maximizes the average performance across shapes.
For each input shape and candidate configuration, we first perform 20 warm-up runs to eliminate cold-start effects, and then run 100 repetitions and take the average as the kernel runtime.
Using SGLang's measured time as the baseline, we report the relative speedups of Astra and our method.
The input shapes are chosen to reflect common dimension configurations in mainstream LLM inference (e.g., typical hidden sizes and head dimensions for LLaMA-7B, 13B, and 70B~\cite{wei2025astra}), ensuring representative evaluation.

\textbf{Model choice.} To reduce confounding factors, ensure fair comparability, and attribute improvements to the proposed closed-loop collaborative mechanism rather than model differences, we use the same base model, deepseek-chat\texttt{(DeepSeek-V3.2)}, across all agents. All experiments are conducted on a single machine equipped with an NVIDIA TITAN RTX.

\section{Results and Analysis}
\subsection{Main Results}

\textbf{Correctness.} All three optimized kernels pass consistency checks against the original SGLang implementations. During tuning, the system performs correctness validation for every candidate configuration to ensure that outputs match the baseline within an acceptable numerical tolerance. The final configurations reported for Astra and our method also satisfy this constraint. These results indicate that our approach preserves operator semantics and introduces no functional deviation, as it only changes execution strategy and resource organization.

\textbf{Performance summary.} Table~\ref{tab:method_compare} reports the speedups of the best general configuration (i.e., the configuration with the best average performance across shapes) relative to the SGL baseline. Astra (multi-agent optimization without templateization) already achieves gains of $1.06\times$--$2.89\times$. By introducing template-based parameterization and search-based autotuning on top of the multi-agent workflow, our method further improves the speedups to $1.09\times$--$3.55\times$, outperforming Astra on all three kernels. Compared with Astra, the additional gain is most pronounced on Kernel1 (about $22.8\%$), while Kernel2/Kernel3 also show consistent improvements (about $2.8\%$/ $4.1\%$). This suggests that templateization does not replace multi-agent optimization; rather, it makes key execution degrees of freedom explicit and tunable, thereby enabling a higher performance ceiling and more stable gains for general configurations.

\begin{table}[h!]
    \centering
    \small
    \caption{Comparison between our method (two-level optimization) and Astra (multi-agent optimization).}
    \label{tab:method_compare}
    \begin{tabular}{lccc}
        \toprule
        Index & SGL (baseline, $\mu\text{s}$) & Astra (Speedup) & Our method (Speedup) \\
        \midrule
        Kernel-1 & 199.15 & 2.89$\times$ & \textbf{3.55$\times$} \\
        Kernel-2 & 163.76 & 1.06$\times$ & \textbf{1.09$\times$} \\
        Kernel-3 & 45.83 & 1.95$\times$ & \textbf{2.03$\times$} \\
        \bottomrule
    \end{tabular}
\end{table}

\subsection{Cross-shape performance of the general configuration}

\begin{figure}[h!]
    \centering
    \includegraphics[width=0.9\textwidth]{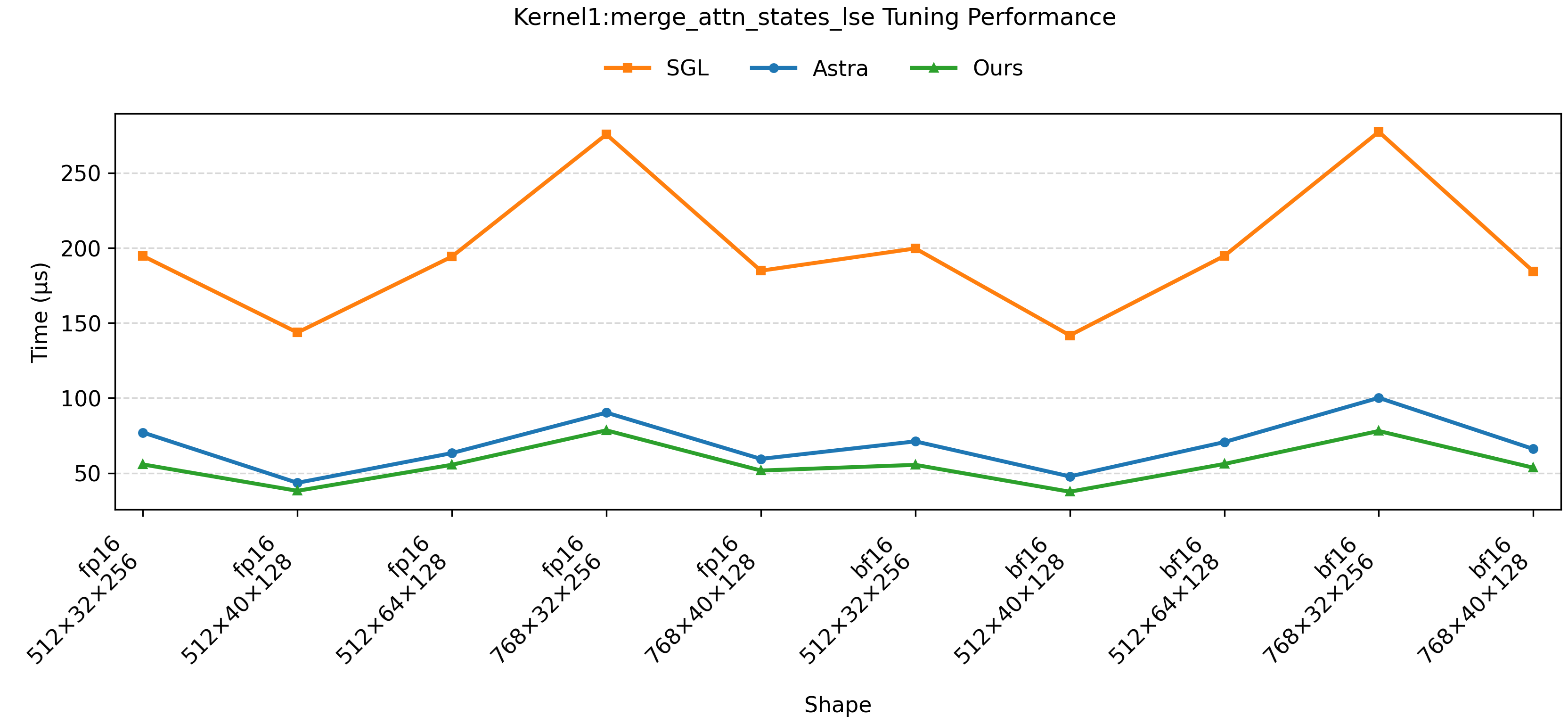}
    \caption{The general configuration that achieves the highest average performance when applied to all shapes of Kernel1. Our method yields lower execution time than Astra and the baseline on most shapes, with a more pronounced advantage on larger shapes.}
    \label{fig:kernel1_perf}
\end{figure}

\begin{figure}[h!]
    \centering
    \includegraphics[width=0.9\textwidth]{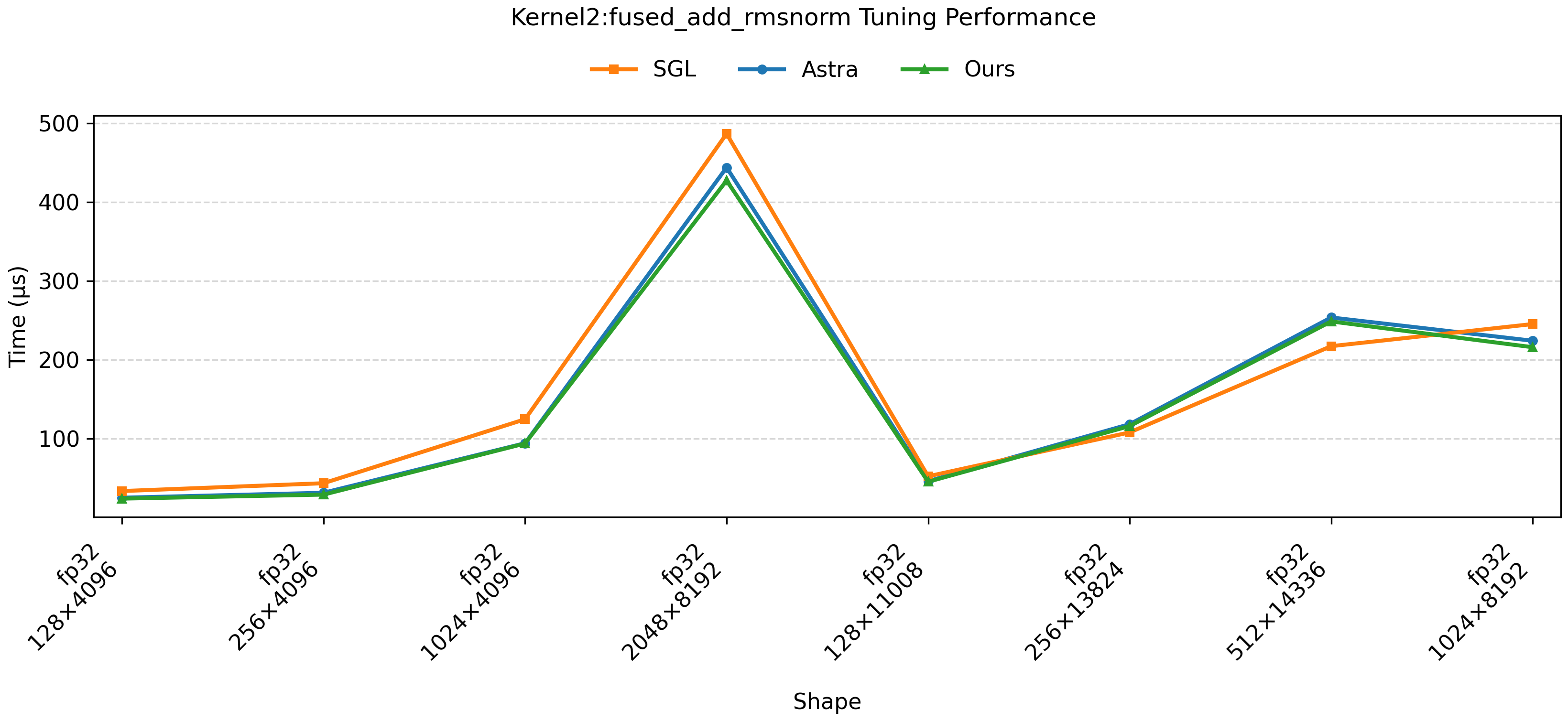}
    \caption{The general configuration that achieves the highest average performance when applied to all shapes of Kernel2. Kernel2 is dominated by problem size: execution time exhibits peak-and-valley behavior across shapes, and our method provides only marginal gains over Astra.}
    \label{fig:kernel2_perf}
\end{figure}

\begin{figure}[h!]
    \centering
    \includegraphics[width=0.9\textwidth]{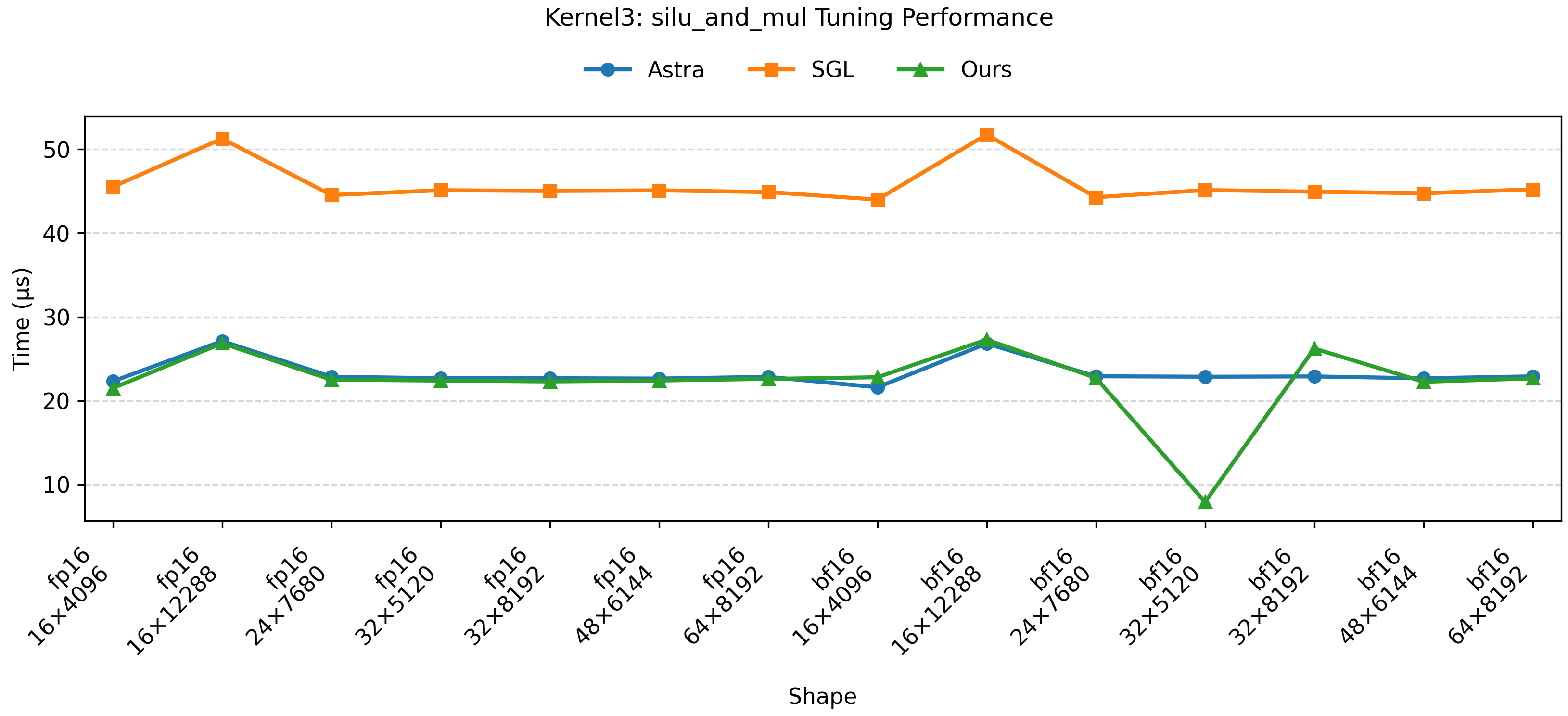}
    \caption{The general configuration that achieves the highest average performance when applied to all shapes of Kernel3.}
    \label{fig:kernel3_perf}
\end{figure}

Figures~\ref{fig:kernel1_perf}--\ref{fig:kernel3_perf} report the average execution time obtained by first selecting, for each kernel, the general configuration that maximizes the average performance across shapes, and then fixing this configuration while evaluating different input shapes. Overall, the three figures exhibit a consistent relative ordering. Compared with the SGLang baseline, Astra substantially reduces execution time across all shapes. Building on this, our method further reduces runtime on most shapes or at least matches Astra, demonstrating that template-based parameterization and search-based autotuning continuously improve the performance ceiling and robustness of the general configuration.

For Kernel1, the gaps among the three curves remain relatively stable across shapes and data types, and our method achieves the lowest execution time for the vast majority of shapes. In particular, for shapes with high baseline cost, the improvement over Astra becomes more evident, indicating that this kernel is more sensitive to execution strategies such as thread-level parallel granularity, loop unrolling, and memory-access organization. Template-based refactoring makes these key degrees of freedom explicit and amenable to systematic search, enabling a single general configuration to remain effective and beneficial in cross-shape settings.

For Kernel2, the runtime varies markedly with shape, showing clear peaks and valleys, with a shared high peak at large shapes, suggesting that performance is primarily dominated by problem size. Both Astra and our method effectively reduce execution time relative to the baseline, while the gap between them is generally small, manifesting as a marginal gain. This indicates that for fused operators whose compute and memory-access structure is relatively fixed and whose dominant bottleneck is clear, multi-agent refactoring already captures most of the available gains, and template-based search mainly serves as fine-grained compensation and parameter matching, yielding robust incremental improvements rather than a structural leap.

For Kernel3, Astra and our method nearly overlap for most shapes, implying that the optimization space under a general-configuration constraint is relatively limited, or that performance is already near the plateau achievable by the chosen parameter family. Nevertheless, for a few shapes, our method still yields further reductions, suggesting that template-based search can unlock additional benefits when alignment, vectorization, or parallel granularity better matches the workload. Overall, our method is never worse than Astra and improves the performance ceiling on specific shapes.

In summary, the results show that multi-agent optimization provides stable baseline gains across shapes, while template-based parameterization and search-based autotuning further improve the transferability and performance ceiling of the general configuration. The improvements are more pronounced for kernels with richer execution degrees of freedom and higher sensitivity to implementation strategies; for kernels with relatively fixed structure or already well-refactored implementations, the effect appears as small but consistent marginal gains, without introducing systematic performance regressions.

\subsection{Shape-sensitivity analysis of performance}

To characterize the attainable performance ceiling of template-based tuning under different problem scales, we further analyze results from the perspective of specialized configurations. Specifically, for each input shape, we select the best parameter configuration for that shape and compare the best achievable performance of different methods across shapes, thereby revealing the sources of performance gains and shape sensitivity.

\begin{figure}[h!]
    \centering
    \includegraphics[width=\textwidth]{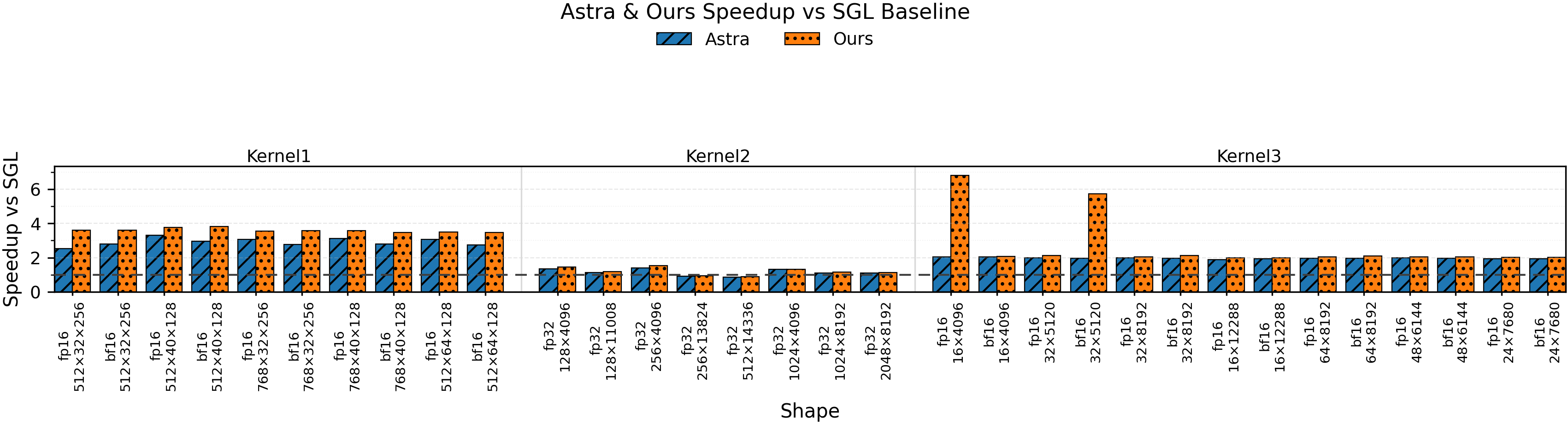}
    \caption{Speedup comparison across input shapes for our method (two-level optimization) versus Astra (multi-agent optimization), relative to the SGLang baseline.}
    \label{fig:speedup_shape}
\end{figure}

In Figure~\ref{fig:speedup_shape}, the dashed line denotes the SGLang baseline. For each kernel, we report the speedup achieved by the shape-specialized best configuration under each input shape, comparing \textbf{Astra} and \textbf{our method}. Overall, both Astra and our method consistently outperform the baseline on the vast majority of shapes, indicating that when shape-specialized configurations are allowed, both approaches can substantially unlock the kernel's performance potential. Meanwhile, our method is better than or comparable to Astra on most shapes, and exhibits clear jumps on a few shapes, demonstrating that template-based parameterization and search can further raise the attainable performance ceiling on top of multi-agent optimization.

Further inspection shows that the best achievable speedup varies substantially across shapes, revealing strong shape sensitivity. The underlying reason is that factors such as alignment and vectorization availability, parallelism and reduction granularity, and the interplay between register pressure and occupancy jointly shift the performance optimum within the parameter space. As a result, a truly shape-agnostic single best configuration rarely exists. Consequently, applying one general configuration to cover multiple shapes inevitably settles at a multi-objective compromise point and degrades average performance; in contrast, shape-specialized configurations more accurately reflect the attainable performance ceiling for each shape.

The results also highlight how structural fitness constrains tuning benefits. When the base implementation structure better matches the parallelism and memory-access characteristics of a given shape, parameter tuning is more likely to yield a higher gain ceiling. Conversely, when structural bottlenecks dominate, tuning parameters alone can hardly deliver improvements of the same magnitude. Based on these observations, we position templateization as a mechanism for making key optimization dimensions explicit and embedding them into a searchable space, enabling evidence-driven adaptive optimization across shapes, rather than attempting to achieve uniformly optimal performance across all shapes with a single general parameter setting.

\section{Conclusion}
This paper presents a controllable automated optimization approach for GPU kernels. Driven by an outer agentic iteration loop, it refactors kernels into templates parameterized by a small set of key execution-strategy dimensions, and performs search-based autotuning under hardware resource constraints with feasibility pruning. In doing so, the approach stably discovers high-quality configurations while preserving operator semantics. Experiments show that the method not only improves peak performance but, more importantly, significantly enhances cross-shape stability and reproducibility. The observed migration of optimal configurations across input shapes confirms strong shape sensitivity, making a single general configuration inevitably a compromise. By exposing a template-based parameter interface and deriving constraints, the system can adapt to different shapes with a controlled tuning budget and provide interpretable sources of speedup. Overall, our results demonstrate that explicit template parameters coupled with hardware-constrained search effectively reduce the randomness of purely generative rewriting, offering a more reliable path toward automated GPU performance optimization for real inference workloads.

\small 
\printbibliography

\normalsize 

\end{document}